\documentclass[lettersize,journal]{IEEEtran}
\usepackage{amsmath,amsfonts}
\usepackage{algorithmic}
\usepackage{algorithm}
\usepackage{array}
\usepackage[caption=false,font=normalsize,labelfont=sf,textfont=sf]{subfig}
\usepackage{textcomp}
\usepackage{stfloats}
\usepackage{url}
\usepackage{verbatim}
\usepackage{graphicx}
\usepackage{cite}
\usepackage{multirow} 
\usepackage{booktabs}
\usepackage{makecell}
\usepackage{xcolor}
\begin{document}

\title{Knowing Where to Focus: Attention-Guided Alignment for Text-based Person Search}

\author{Lei Tan, Weihao Li, Pingyang Dai, Jie Chen~\IEEEmembership{Member,~IEEE}, \\
Liujuan Cao~\IEEEmembership{Member,~IEEE}, Rongrong Ji~\IEEEmembership{Senior Member,~IEEE}
\thanks{Lei Tan, Weihao Li, Pingyang Dai, Liujuan Cao, and Rongrong Ji are with the Media Analytics and Computing Laboratory, Department of Artificial Intelligence, School of Informatics, Xiamen University, Xiamen 361005, China (e-mail:
tanlei@stu.xmu.edu.cn; liweihao@stu.xmu.edu.cn; pydai@xmu.edu.cn; caoliujuan@xmu.edu.cn; rrji@xmu.edu.cn).}
\thanks{Jie Chen is with the School of Electronic and Computer Engineering, Peking University, Beijing 100871, China, and also with Peng Cheng Laboratory,
Shenzhen 518066, China (e-mail: chenj@pcl.ac.cn).}}



\maketitle

\begin{abstract}
In the realm of Text-Based Person Search (TBPS), mainstream methods aim to explore more efficient interaction frameworks between text descriptions and visual data. However, recent approaches encounter two principal challenges. Firstly, the widely used random-based Masked Language Modeling (MLM) considers all the words in the text equally during training. However, massive semantically vacuous words ('with', 'the', etc.) be masked fail to contribute efficient interaction in the cross-modal MLM and hampers the representation alignment. Secondly, manual descriptions in TBPS datasets are tedious and inevitably contain several inaccuracies. To address these issues, we introduce an Attention-Guided Alignment (AGA) framework featuring two innovative components: Attention-Guided Mask (AGM) Modeling and Text Enrichment Module (TEM). AGM dynamically masks semantically meaningful words by aggregating the attention weight derived from the text encoding process, thereby cross-modal MLM can capture information related to the masked word from text context and images and align their representations. Meanwhile, TEM alleviates low-quality representations caused by repetitive and erroneous text descriptions by replacing those semantically meaningful words with MLM's prediction. It not only enriches text descriptions but also prevents overfitting. Extensive experiments across three challenging benchmarks demonstrate the effectiveness of our AGA, achieving new state-of-the-art results with Rank-1 accuracy reaching \(78.36\%\), \(67.31\%\), and \(67.4\%\) on CUHK-PEDES, ICFG-PEDES, and RSTPReid, respectively.
\end{abstract}


\section{Introduction}
\label{sec:intro}
Text-based Person Search (TBPS) aims to retrieve images of a target person from an image gallery using natural language descriptions as query input. Traditional image person re-identification~\cite{he2021transreid,tan2024partformer,tan2024occluded} assumes that an image of the target person can be obtained as the query during the retrieval processing. However, the query image is not always available in some cases. Compared to images-based person search~\cite{wang2022body, tan2022dynamic,zhang2022person,wei2023dual,gu2022multi,han2024self,he2023region}, textual descriptions of a person are easier to obtain and fulfill practical application requirements. Consequently, TBPS has gained significant attention in recent years and has found extensive applications in smart cities and security surveillance.

\begin{figure}[t]
\centering
\includegraphics[width=0.95\columnwidth]{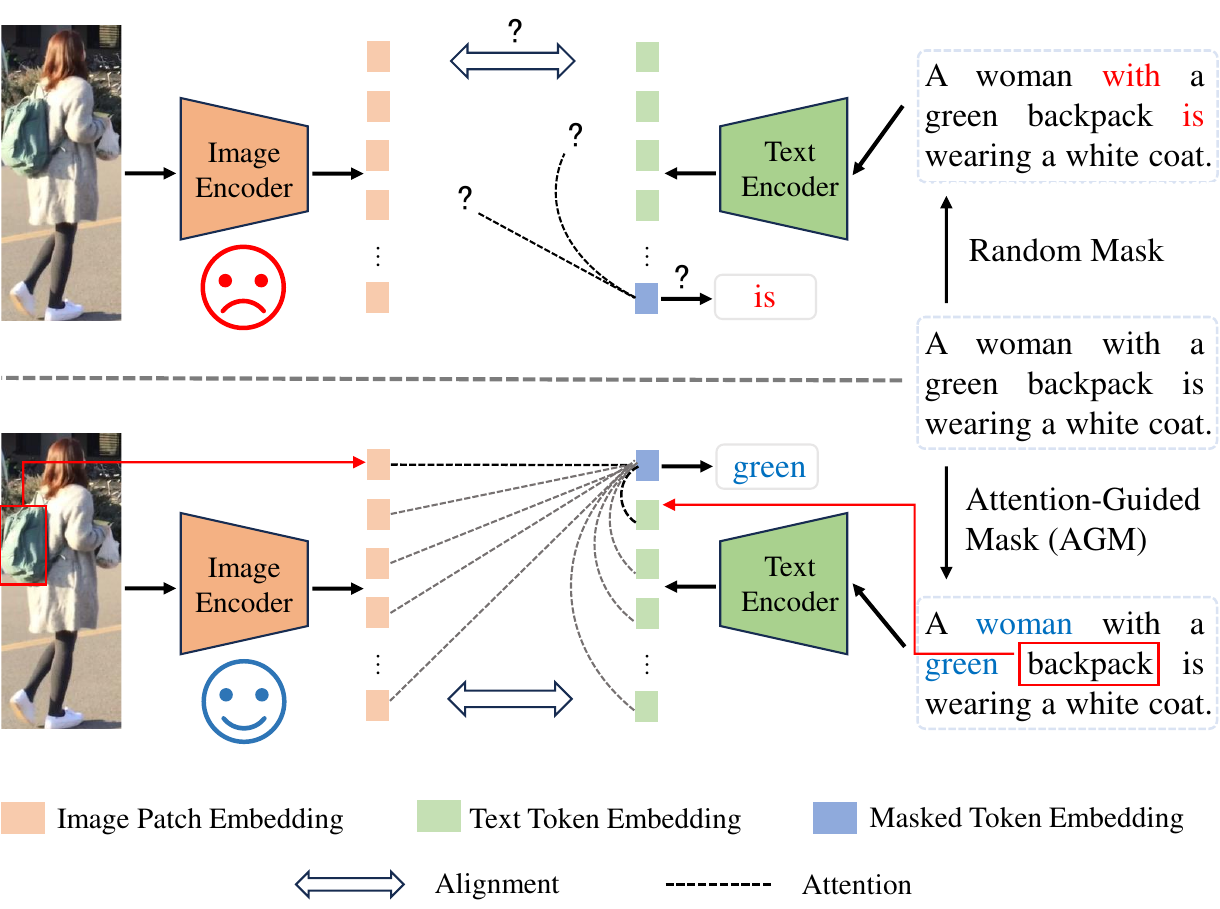}
\caption{\label{fig}Illustration of the motivation of our AGM. The upper part depicts the traditional random mask strategy which easily masks meaningless words, making it impossible to align the corresponding semantics of text context and image. The lower part is the proposed AGM strategy, which can locate semantically meaningful words, thus facilitating cross-modal alignment.}
\end{figure}
A primary challenge within TBPS is to seek alignment between multi-modal feature representations. Recently, Visual-Language Pre-training (VLP) models have made rapid advancements, offering new solutions for downstream TBPS tasks. Trained on a massive dataset of text-image pairs, VLP models exhibit superior semantic understanding and cross-modal alignment capabilities. Some works~\cite{han2021textreid, IRRA, bai2023rasa} have attempted to adopt the VLP models, such as CLIP~\cite{clip}, ALBEF~\cite{albef}, in downstream TBPS task, yielding promising results. During this processing, cross-modal Masked Language Modeling (MLM) is been demonstrated as an efficient way for cross-modal feature alignment~\cite{albef} and has been widely used in TBPS~\cite{IRRA, bai2023rasa}. 
Cross-modal MLM aims to utilize both visual and textual context information to predict masked words and improve the interaction between text descriptions and images. During the prediction of original words, the attention mechanism captures the semantic information related to the masked words from the image and text context, thereby enhancing semantic alignment across modalities. However, it is worth noting that when VLP models are used in downstream TBPS tasks, cross-modal MLM aims to align image and text context semantics representation using specific masked tokens as pivots, rather than remodeling the entire representational space as BERT does. Such a variation reduces the difficulty of learning and ensures performance under limited TBPS data.
However, current methods adopt a uniform random masking strategy to indiscriminately mask words, in which semantically vacuous words like 'with' or 'the' show the same probability of being masked as those semantically meaningful words. 
These semantically vacuous words lack relevant details in the image and text context, causing failure to capture corresponding background information (image and text context) and align their representations when predicting the original masked words. 
A piece of strong evidence is illustrated by a toy experiment. As shown in Table~\ref{table:toy}, masks on those semantically vacuous words largely limited the performance of MLM strategy.
Inspired by this observation, we attempt to encourage the MLM to pay more attention to semantically meaningful words. It is worth noticing that the attention results of the class token in the text encoder, which is generated through a process of weighted summation, can serve as an indicator of the importance of each text token. Therefore, we introduce the Attention-Guided Mask (AGM) Modeling module, leveraging textual class attention weight to selectively semantically meaningful words. 
Unlike static manual or knowledge-based masking rules, AGM is a novel way that can dynamically mask words for individual inputs. Meanwhile, AGM does not impose any additional computational overhead compared to random masking strategies because the attention weights are conveniently obtained during the forward encoding processing of descriptions.
\begin{table}[t]
\small
\renewcommand\arraystretch{1.2}
\setlength{\tabcolsep}{5pt}
\centering
\caption{Comparison of different proportions of the vacuous words on CUHK-PEDES. $Ratio_{v}$ is defined as $N_{vacuous}/N_{Masked}$. For the 'Picked' setting, we created a small vocabulary containing common semantically vacuous words like 'the', 'and', 'that', etc. 'Picked' only masks meaningless words. 'Random' adopts a random mask strategy. 'Baseline' does not do MLM. Since calculating all the sentences in the dataset needs a great effort, we only counted the first 100 sentences to calculate the $Ratio_{v}$. Clearly, AGM decreases the masking probability of meaningless words, thus improving the final performance.}
\resizebox{0.9\columnwidth}{!}
{\begin{tabular}{c|c|cccc} 
\hline
Method      &$Ratio_{v}$     & R@1            & R@5            & R@10           & mAP             \\ 
\hline
Baseline        &-           & 76.03          & 90.31          & 93.73          & 69.23       \\
Picked        &100\%           & 75.50          & 89.24          & 93.63          & 69.15       \\
Random    &$\approx$50\%            & 76.51          & 90.29          & 94.25          & 69.38           \\
AGM      &$\approx$8.3\%  & \textbf{77.58} & \textbf{91.61} & \textbf{94.89} & \textbf{70.50}  \\
\hline
\end{tabular}}
\label{table:toy}
\end{table}

Although MLM makes sense in the TBPS and reaches passable performance. Due to the quality of text descriptions in the downstream TBPS datasets, the training usually tends to be sub-optimal. As illustrated by APTM~\cite{zhedong}, the manually annotated text description is not accurate enough and the words are rather common. For example, 'women' is used frequently, while 'lady' is rarely seen in the dataset, and as shown in Fig~\ref{fig}, when describing the color of pants, 'dark' or 'gray' can describe the image more accurately than the original 'black', these problems potentially undermines the rich representational power of VLP models and can easily lead to overfitting. 
To address this problem, we attempt the automatically revise the text description in a self-supervised manner.
Since predicting the mask words is a classifying task, the output logits can reflect the probability of a token belonging to each word in the vocabulary.
Due to the VLP's powerful knowledge and the helping of image information, the heightened logits words show a strong association with the image, and the top logits words can even better describe the specific content of the image, for example, in Figure~\ref{fig}, the description for the color of the pedestrian's trousers could be more accurately revised from 'black' to 'dark' or 'gray', and these words indeed appear among the top logits words. Inspired by the above observation, we further proposed a Text Enrichment Module (TEM). 
When AGM adeptly identifies words rich in semantic content for masking, we further use TEM to augment the original descriptions with its logit-based alternatives. TEM makes text descriptions more accurate and diverse, which retains powerful knowledge of VLP and effectively inhibits VLP from overfitting on the downstream TBPS datasets. Following~\cite{albef}, we also conducted a simple theoretical proof of this method based on the perspective of mutual information maximization.
The main contributions of this paper are summarized as follows:

\begin{itemize}
    \item We propose an Attention-Guided Alignment (AGA) framework for text-based person search to encourage a more stable and efficient interactive training between the text descriptions and image samples.
    \item We introduce a novel Attention-Guided Mask (AGM) Modeling together with the Text Enrichment Module (TEM). The former targets the masking of semantically meaningful words to facilitate the training of cross-modal alignment, while the latter ensures the accuracy and diversity of text descriptions, thereby preventing the model from overfitting and degradation.
    \item Extensive experiments demonstrate the effectiveness of our approach, which outperforms existing state-of-the-art methods, achieving Rank-1 accuracies of \(78.36\%\), \(67.31\%\), and \(67.4\%\) on CUHK-PEDES, ICFG-PEDES, and RSTPReid datasets, respectively.
\end{itemize}

\section{Related Work}
\label{sec:RW}
Text-based Person Search (TBPS) is a cross-modal retrieval task that was first proposed by Li et al.~\cite{li2017person}, who also published the first challenging benchmark dataset CUHK-PEDES. Early approaches~\cite{zhang2018deep, ding2021semantically} primarily employed an unimodal encoder, such as ResNet~\cite{resnet} and Bert~\cite{bert}, to separately extract features from text and image. Subsequently, these features were projected onto a shared latent space using a projection layer to achieve global or local feature alignment, CMPM/C~\cite{zhang2018deep} proposes a cross-modal projection matching (CMPM) loss and a cross-modal projection classification (CMPC) for learning more discriminative image-text embeddings. SSAN~\cite{ding2021semantically} tries to align visual and textual part features, such as body parts and text phrases, Moreover, ACSA~\cite{ji2022asymmetric} proposes an asymmetric alignment scheme, this work found a word (such as 'tall', 'thin', etc.) may correspond to the whole person. Recently, large-scale Vision-Language Pre-training (VLP) models, such as CLIP~\cite{clip} and ALBEF~\cite{albef}, have provided new solutions for the TBPS task. PSLD~\cite{han2021textreid} applies momentum contrastive learning to CLIP to learn better latent space feature representations on smaller TBPS datasets. CFine~\cite{yan2023clip} proposes a multi-modal interaction module based on CLIP to mine fine-grained information of image-text pairs. IRRA~\cite{IRRA} solved the problem of without modal interaction in CLIP during pre-training and found cross-modal MLM that utilizes the image and text contextualized information to predict masked word can better align image and text contextualized representations. On its opposite, BiLMa~\cite{fujii2023bilma} masks image patches to perform cross-modal MLM and also achieves good results. Rasa~\cite{bai2023rasa} proposed a Relation-aware Learning module and Sensitivity-aware Learning module, the former allows the model to better utilize weak positive samples, and the latter allows the model to learn more fine-grained information by guiding the model to perceive whether the word in the text is replaced. MLLM-ReID~\cite{tan2024harnessing} leverages Multi-modal Large Language Models (MLLM) to generate high-quality, diverse descriptions to improve the quality of datasets. PLOT~\cite{park2024plot} leverages a part discovery module based on slot attention to autonomously identify and align distinctive parts across modalities, enhancing interpretability and retrieval accuracy without explicit part-level correspondence supervision. Although these methods have achieved remarkable improvement for TBPS in recent years, they still struggle with the alignment problem between the text and image content.   

\subsection{Masked Language Modeling}
Taylor~\cite{taylor} proposed Masked Language Modeling (MLM) in 1953, it became widely known when BERT~\cite{bert} used it as a pre-training task for language model training. It replaces some words in the text with a special token [MASK], and the output of the last Transformer block corresponding to the mask tokens will fed into a softmax layer to predict the original words by utilizing contextual information of the text. This autoregressive training approach allows massive amounts of text on the internet to be used for model training, which also propels the rapid development of NLP. Recently, ALBEF~\cite{albef} found MLM also can be used for Vision-Language Pre-training, it aims to predict masked textual tokens not only by the rest of the unmasked textual tokens but also using the visual tokens, the masked textual tokens serve as an anchor to align semantic representation of visual and textual contextual information. IRRA~\cite{IRRA} use a similar cross-modal MLM method on the downstream TBPS task can further align cross-modal features, supplementing the fact that CLIP has no interaction between text and image information during pre-training, which achieved good performance. Rasa~\cite{bai2023rasa} uses ALBEF~\cite{albef} as the backbone and keeps the cross-modal MLM method setting during training on the downstream TBPS task. Although recent MLM strategies have shown the ability to align images with text, the randomized text selection makes it difficult to fully unleash its power.

\begin{figure*}[t]
  \centerline{\includegraphics[width=0.95\textwidth]{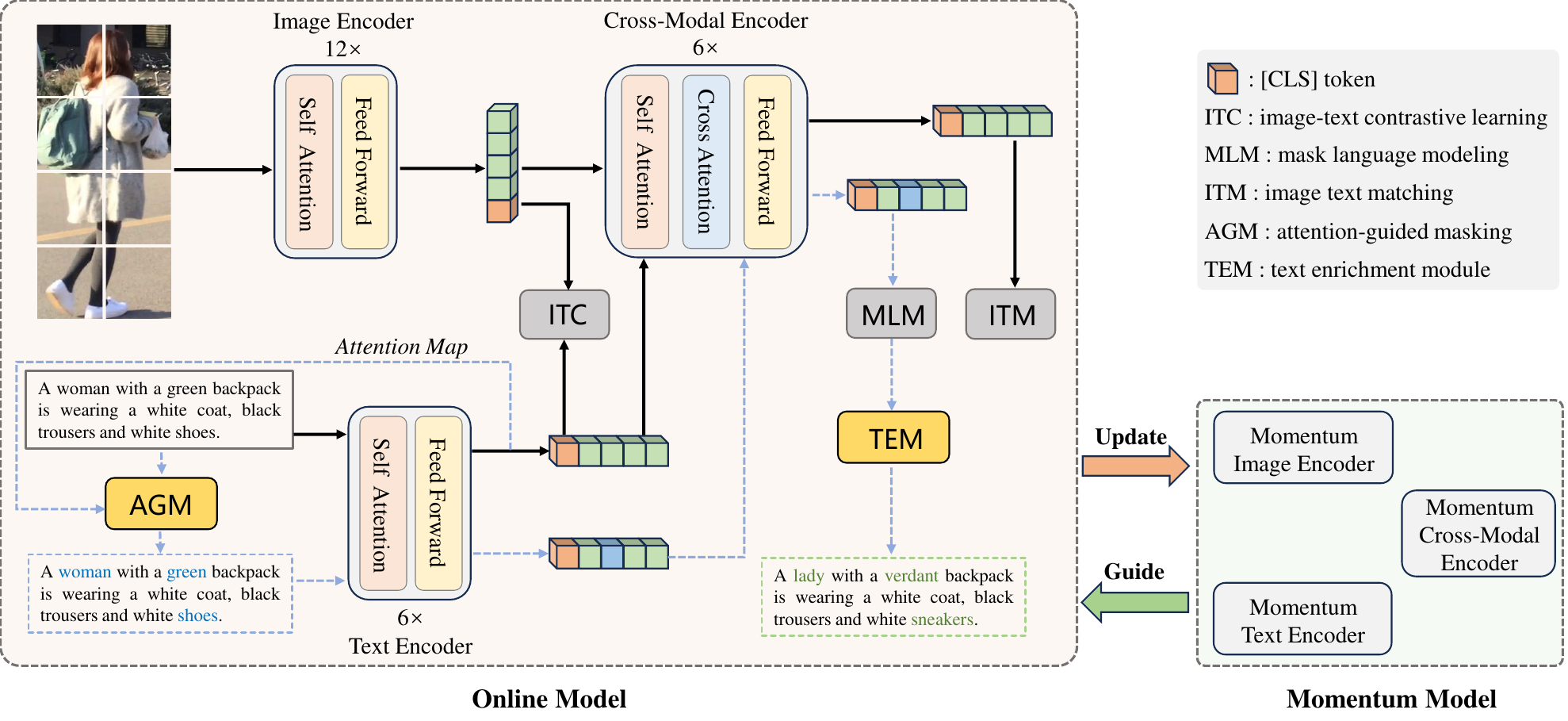}}
  \caption{\textbf{Model architecture of our method.} AGA consists of an image encoder, a text encoder, and a cross-modal encoder. AGM aims to select meaningful words for masking by referring to class attention. TEM replaces the original words based on the logit of the MLM head, thereby enriching the original text description. The momentum model (a slow-moving of the online model) is used to guide the online model to learn better representations.}
  \label{fig2}
\end{figure*}

\section{Method}
\label{sec:method}
To encourage a more stable and efficient interactive training between the text descriptions and image samples, we propose an Attention-Guided Alignment (AGA) framework for text-based person search. Therefore, in this section, we first provide a brief overview of the basic structure of AGA, and then highlight the proposed Attention Guided Mask (AGM) Modeling and Text Enrichment Module (TEM) in detail.

\subsection{Model Architecture}
\label{sec:3.1}
As illustrated in Figure~\ref{fig2}, following the success of ALBEF~\cite{albef},  the proposed Attention-Guided Alignment (AGA) framework consists of an online model and a momentum model with the same model architecture. 
The momentum model, serving as a smoothed and stabilized version of the online model, is maintained through an Exponential Moving Average (EMA) of the online model's parameters. Encoded feature from the momentum model is then employed as pseudo labels to guide the representational learning of the online model.

The architecture of the model includes a 12-layer Vision Transformer (ViT-B/16) for image encoding and two 6-layer transformer blocks for text encoding and cross-modal interaction. The weight of the text encoder and cross-modal encoder are initialized from the first 6 layers and last 6 layers of the pre-trained BERT~\cite{bert} respectively. Given an image-text pair$(I, T)$, we feed the image $I$ and text $T$ into the corresponding encoder to obtain the visual embedding $\{v_{cls}, v_1,\cdots,v_M\}$ and textual embedding $\{t_{cls}, t_1,\cdots,t_N\}$, $M$ and $N$ is the number of images patches and text tokens respectively,  $v_{cls}$ and $t_{cls}$ as the global representation and then feed to compute image-text contrastive learning (ITC) loss, specifically, for text, the softmax-normalized text-to-image similarity is computed as:
\begin{equation}
\label{eqn:sim}
p_m^\mathrm{t2i}(T) = \frac{\exp (s(t_{cls},I_m)/ \tau)}{\sum_{m=1}^M \exp (s(t_{cls},I_m)/ \tau)},
\end{equation}
where $\tau$ represents a temperature parameter and $s(\cdot)$ denotes the similarity function. Analogously, image-to-text similarity $p_m^{\mathrm{i2t}}$ is computed likewise, and the comprehensive ITC loss is denoted as:
\begin{align}
\label{eqn:itc}
\mathcal{L}_\mathrm{itc}  = \frac{1}{2} \mathbb{E}_{(I,T)\sim D} &\big[ \mathrm{H}({y}^\mathrm{i2t}(I),{p}^\mathrm{i2t}(I)) \nonumber\\
& +\mathrm{H}({y}^\mathrm{t2i}(T),{p}^\mathrm{t2i}(T)) \big],
\end{align}  
where the $y^{t2i}$ and $y^{i2t}$ denote the ground-truth one-hot labels. It is worth mentioning that momentum distillation image-text contrastive learning is used in ALBEF, taking text as an example, the corresponding positive image will be encoded by the momentum model and then the obtained features also will be sent to the queue as positive samples during contrastive learning.

The input of the cross-modal encoder includes textual embedding and visual embedding, the former is as a query and later serves as key and value. The last hidden state of the cross-modal encoder can be denoted as $\{f_{cls}, f_1,\cdots,f_N\}$, $f_{cls}$ will use to compute ITM loss as Eq.~\eqref{eqn:itm}, and masked token in $\{f_1,\cdots,f_N\}$ fed to MLM head to compute MLM loss as Eq.~\eqref{eqn:mlm}:
\begin{equation}
\label{eqn:itm}
\mathcal{L}_\mathrm{itm} = \mathbb{E}_{(I,T)\sim D} \mathrm{H} ( {y}^\textrm{itm},  {p}^\textrm{itm}(I,T)),
\end{equation}
\begin{equation}
\label{eqn:mlm}
\mathcal{L}_\mathrm{mlm} = \mathbb{E}_{(I,\hat{T})\sim D} \mathrm{H} ( {y}^\textrm{msk},  {p}^\textrm{msk}(I,\hat{T})).
\end{equation}
RaSa~\cite{bai2023rasa} uses ALBEF as the backbone and further proposes the Ra module and Sa module, the former aimed at effectively harnessing weak positive text to supplement information while mitigating the impact of noise. The Sa module discerns word replacements in sentence descriptions, sharpening the model's perception of pertinent information. Our method follows these settings.

\begin{figure}[t]
\centering
\includegraphics[width=0.95\columnwidth]{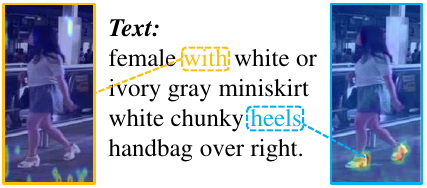}
\caption{\label{fig:3} Visualization results of cross attention map of masked words. \textbf{Left} masked the word 'with' while \textbf{(Right)} masked the word 'heels'.}
\end{figure}

\subsection{Attention-Guided Mask Modeling}
\label{sec:3.2}
\textbf{Limitation of vanilla MLM.} The Masked Language Modeling (MLM) aims to predict a masked subset of input tokens based on the remaining context. The vanilla MLM adopts a random mask strategy, in which each word has a 15\% probability of being selected, among the selected words, 80\% was replaced with [MASK] token, 10\% was randomly replaced with other words, and 10\% was unchanged. After that, the MLM works toward the cross-modal encoder which consists of a multi-head self-attention layer (MHSA), a multi-head cross-attention (MHCA) layer, and a feed-forward network (FFN). Since the MHSA and FFN do not contain the interaction between the two modalities, therefore we only consider the MHCA here. Let $\hat{T}$ denotes a masked text, the processing of MHCA is shown as:
\begin{equation}
\label{eq:attn}
Attn(Q_{l,\hat{T}},K_{l,I},V_{l,I})=\sigma(\frac{Q_{l,\hat{T}}(K_{l,i})^T}{\sqrt{d}})V_{l,i},\\
\end{equation}
where the $Q_{l,\hat{T}}$ refers to the query matrix generated by the masked text in the $l_{th}$ cross-modal encoder, $K_{l,I}$ and $V_{l,I}$ refers to the key and value matrix generated by the image tokens in the $l_{th}$ cross-modal encoder. Then the masked tokens in the last latent status will feed to the MLM head to predict the original words, which utilize the remaining text context tokens and image tokens, those masked tokens as anchors, capturing semantic information related to those masked tokens from text context and images through the attention process to achieve the purpose of cross-modal representation alignment. For example, as shown in Figure~\ref{fig:3}, when the masked word is 'heels', 'white chunk' in the text and relevant patches in the image must be captured in order to predict the original word 'heels'. However, the vanilla cross-modal MLM all use random mask strategy, which leads to a large portion of the masked words being uninformative like 'the,' 'also,' 'with,' etc. These words have no relevant semantics in the text context and image so it is impossible to align feature representations across modal. The visualization of the cross-attention map between masked tokens and image patches reflects this problem. When using VLP for downstream tasks, the MLM method aims to align cross-modal representations, rather than modeling the entire representation space like BERT. Masking these uninformative words is not only unhelpful for training, but may even destroy the representation space.

\textbf{Attention-Guided Mask Modeling.} Regarding the issue above, we propose a novel Attention-Guided Mask (AGM) Modeling to fully explore the potential of cross-modal MLM. Compared to the vanilla MLM which uses the equally mask probability to all the words, the AGM employs a dynamic probability instead. As is known to all, the attention weight of class token in the transformer blocks can reflect the importance of each token, the side reflects whether the word is informative, and it naturally be able to used as a reference for the probability of the word being masked. Therefore, in AGM, we explore the potential of using the guidance of the class attention weight of the text to dynamically adjust the probability of words being masked. Specifically, for the text $T$, we can formulate its class attention weight in each layer as:
\begin{equation}
\label{eq:attn2}
A^{CLS} = [\mathbf{a}^{cls}_0, \mathbf{a}^{cls}_1, \mathbf{a}^{cls}_2, \dots ,\mathbf{a}^{cls}_K  ]
\end{equation}
where $a^{cls}_k$ refers to the class attention weight in the $k_{th}$ layer, and $K$ is the number of layers in the text encoder. The most intuitive strategy is to average each layer to obtain the final class attention weight, but as the network goes deeper, the noise will be decreased, and the network can focus more on the regions with rich semantics, those attention weights in the deep layer will be more valuable compared to the shallow layers. Therefore, in our AGM, we take the combined class attention weight across different transformer layers by using the exponential moving average as:
\begin{align}
\label{ema}
\bar{\mathbf{a}}_{k} = \beta \cdot \bar{\mathbf{a}}_{k-1} + (1-\beta) \cdot {a}_{k}^{cls}, 
\end{align}
where we set $\beta = 0.95$, $k \in \{1, 2, \ldots, 6\}$, $\bar{\mathbf{a}}_0=0$, and $\bar{\mathbf{a}}_{6}$ as the final class attention weight. It is worth noting that $\bar{\mathbf{a}}_{6}$ contain a class token self-attention weight, but the class token does not correspond to specific words, so we reassign the class attention weight as Eq.~\eqref{soft} to ignore the influence of the class token self-attention as: 
\begin{align}
\label{soft}
\bar{a}_i = \frac{e^{(\bar{a}_i/\tau)}}{\sum_{j=1}^{n} e^{(\bar{a}_j/\tau)}}, i \in \left[1, n\right],
\end{align}
where $\bar{a}_i$ is denoted the class attention weight of $i_{th}$ token and $\tau$ is the temperature parameter which dynamically adjusts the range and intervals of the softmax results, we set it as 0.02, and $n$ is the number of tokens. Then we adopt the value after softmax as a reference to mask the corresponding word, specifically, the probability of masking $i_{th}$ token is denoted as:
\begin{align}\label{prob}
p_i = \alpha_1 + \alpha_2 \times {\bar{a}_i}
\end{align}
where $\alpha_1$ is offset which can also be considered as a lower bound of masking probability to each text, and $\alpha_2$ is the amplitude, both of them can be set to control the ratio of words being masked in the entire text.


\subsection{Text Enrichment Module}

\label{sec:3.3}
\begin{figure}[t]
\centering
\includegraphics[width=0.95\columnwidth]{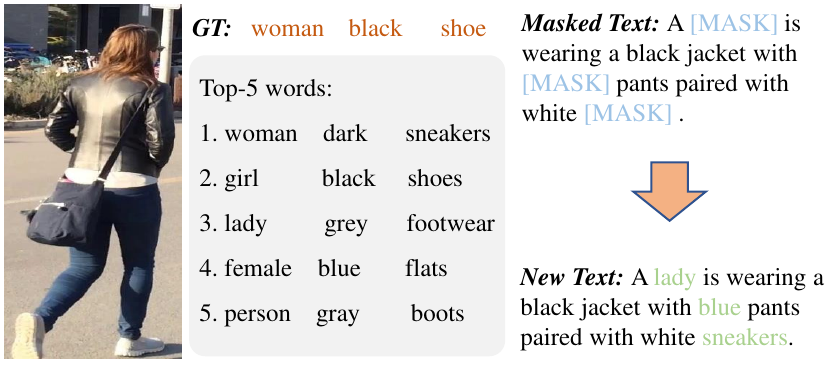}
\caption{\label{fig:TEM}Example of Text Enrichment Module (TEM). TEM enhances textual descriptions by replacing original words based on the logit of the MLM head, resulting in richer and more precise descriptions.}
\end{figure}

The manual description is tedious and inevitably introduces several inaccurate. As shown in Figure~\ref{fig:TEM}, the jeans described as 'black' are actually 'blue'. On the other hand, the description of the same object is tedious, for example, 'shoe' is more commonly used, and 'sneaker' is seldom seen in descriptions. Due to the limitation of the data scale and various annotations with human bias, VLP is prone to overfitting and losing powerful knowledge when trained on the downstream TBPS datasets. To tackle this issue, based on the AGM which can mask words to more informative ones, we further propose a Text Enrichment Module (TEM) to provide diversity and high-quality text descriptions. Due to VLP has strong semantic understanding capabilities, the logits of MLM head can reflect the semantic similarity between the predicted word and the original masked word, and benefit from MLM utilizing image information when predicting original words, these words with high logit can even more accurately describe the corresponding content in the image. As shown in Figure~\ref{fig:TEM}, we show the Top-5 predicted words and the original ground-truth words, it is clear that all the words seem suitable to describe the content in the image and even better, for example, 'dark' may be more appropriate to describe the color of pants than ground-truth 'black'. Following this idea, we attempt to explore a novel self-supervised manner to avoid losing the powerful ability of VLP. Specifically, we replace the original masked words with reference logits by a probability $p_{tem}$, aiming to enrich the original text description and decrease the noises without additional resource consumption. For a masked text $\hat{T}$ and its prediction logits $p_T$, the potential top list is given as:
\begin{equation}\label{eq:pt}
\bar{p}_T = \sigma(R^m(p_T)),
\end{equation}
where $\sigma$ refers to the softmax function and the $R^m(\cdot)$ indicate the top-m logits. Then we get a potential replacement list for the masked text $\hat{T}$, we randomly sample a word from it, and the sampled word is different from the original word, this process can be formalized as:
\begin{equation}\label{eq:sample}
W_i \sim \operatorname{Multinomial}(1, \bar{p}_T).
\end{equation}
Then we replace original words with new words to obtain a new text description, replacement of meaningful words with synonyms, not only preserves the original information but is also even more accurate for the corresponding picture, this is due to the fact that not only text context but also picture information is used to predict masked words. Then we replaced the original sentence with the new sentence with probability $p_{tem}$.

Here, we also provide a simple analysis of TEM. Reviewing ALBEF's momentum distillation ITC, based on Eq.~\eqref{eqn:itm}, an additional distillation loss is added. the momentum distillation loss is defined as:
\begin{align}\label{eq:kl}
\mathcal{L}_\mathrm{itc}^\mathrm{mod} = \mathbb{E}_{(I,T)\sim D} &\big[ \mathrm{KL}({q}^\mathrm{i2t}(I) \parallel {p}^\mathrm{i2t}(I)) \nonumber \\
&+ \mathrm{KL}({q}^\mathrm{t2i}(T)\parallel{p}^\mathrm{t2i}(T))\big]
\end{align}
As demonstrated in ALBEF, we also interpret TEM as maximizing a lower bound on the mutual information (MI) between different 'views' of an image-text pair. We define two random variables $a$ and $b$ as features encoded by online models and momentum models. ALBEF treats $a$ and $b$ as different views of the data point, maximizing the lower bound of mutual information by optimizing the KL distance between a and b as Eq.~\eqref{eq:kl}, which aims to learn representations invariant to the change of view. From the perspective of text semantics, our TEM uses a clever way to replace the original words, obtaining a new text description that still retains the same semantics of the original text and further creates a new 'view' of the text.


\section{Experiments}
\begin{table*}[t]
\small
\centering
\caption{Comparison with other methods on CUHK-PEDES, ICFG-PEDES, and RSTPReid. w/o VLP means without taking the VLP model as the backbone. For a fair comparison, all reported results come from the methods without re-ranking.}
\tabcolsep=2pt
\renewcommand\arraystretch{1.2}
\resizebox{1.95\columnwidth}{!}{
\begin{tabular}{cc|c|cccc|cccc|cccc}
\hline
\multicolumn{2}{c|}{\multirow{2}{*}{\textbf{Method}}} & \multirow{2}{*}{\textbf{Venus}}  & \multicolumn{4}{c|}{\textbf{CUHK-PEDES}}         & \multicolumn{4}{c|}{\textbf{ICFG-PEDES}}           & \multicolumn{4}{c}{\textbf{RSTPReid}} \\
&    &    & R@1  & R@5  & R@10   & mAP              & R@1  & R@5  & R@10   & mAP                & R@1  & R@5  & R@10  & mAP \\
\hline
\multirow{12}{*}{\rotatebox{90}{w/o VLP}}      & Dual Path \cite{zheng2020dual}   & TOMM'20 & 44.40     & 66.26      & 75.07      & -        & 38.99     & 59.44     & 68.41      & -         & -        & -         & -         & -     \\
                                               & CMPM/C \cite{zhang2018deep}      & ECCV'18 & 49.37     & 71.69      & 79.27      & -        & 43.51     & 65.44     & 74.26      & -         & -        & -         & -         & -     \\
                                               & ViTAA \cite{wang2020vitaa}       & ECCV'20 & 55.97     & 75.84      & 83.52      & -        & 50.98     & 68.79     & 75.78      & -         & -        & -         & -         & -        \\
                                               & DSSL \cite{zhu2021dssl}          & MM'21     & 80.41      & 87.56      & -        & -         & -         & -          & -         & 32.43    & 55.08     & 63.19     & - \\
                                               & MGEL \cite{wang2021text}         & IJCAI'21     & 60.27     & 80.01      & 86.74      & -        & -         & -         & -          & -         & -        & -         & -         & -         \\
                                               & SSAN \cite{ding2021semantically} & Arxiv'21     & 61.37     & 80.15      & 86.73      & -        & 54.23     & 72.63     & 79.53      & -         & 43.50    & 67.80     & 77.15     & - \\
                                               & ACSA \cite{ji2022asymmetric}     & TMM'22     & 63.56     & 81.40      & 87.70      & -        & -         & -         & -          & -         & 48.40    & 71.85     & 81.45     & -\\
                                               & SAF \cite{li2022learning}        & ICASSP'21     & 64.13     & 82.62      & 88.40      & 58.61    & 54.86     & 72.13     & 79.13      & 32.76     & 44.05    & 67.30     & 76.25     & 36.81\\
                                               & TIPCB \cite{chen2022tipcb}       & NC'22     & 64.26     & 83.19      & 89.10      & -        & 54.96     & 74.72     & 81.89      & -         & -        & -         & -         & -    \\
                                               & CAIBC \cite{wang2022caibc}       & MM'22     & 64.43     & 82.87      & 88.37      & -        & -         & -         & -          & -         & 47.35    & 69.55     & 79.00     & -\\
                                               & $\rm C_2A_2$ \cite{niu2022cross} & MM'22     & 64.82     & 83.54      & 89.77      & -        & -         & -         & -          & -         & 51.55    & 76.75     & 85.15     & -\\
                                               & LGUR \cite{shao2022learning}     & MM'22     & 65.25     & 83.12      & 89.00      & -        & 59.02     & 75.32     & 81.56      & -         & -        & -         & -         & -      \\
\hline\hline
\multirow{5}{*}{\rotatebox{90}{w/ VLP}}        & PSLD \cite{han2021textreid}      & BMVC'22     & 64.08     & 81.73      & 88.19      & 60.08    & -         & -         & -          & -         & -        & -         & -         & -       \\
                                               & IVT \cite{shu2022see}            & ECCV'22     & 65.59     & 83.11      & 89.21      & -        & 56.04     & 73.60     & 80.22      & -         & 46.70    & 70.00     & 78.80     & -\\
                                               & CFine \cite{yan2023clip}         & TIP'23    & 69.57     & 85.93      & 91.15      & -        & 60.83     & 76.55     & 82.42      & -         & 50.55    & 72.50     & 81.60     & -\\
                                               & IRRA \cite{IRRA}                 & CVPR'23    & 73.38     & 89.93     & 93.71       & 66.13    & 63.46     & 80.25     & 85.82      & 38.06     & 60.2     & 81.30      & 88.2      & 47.17\\
                                               & BiLMa \cite{fujii2023bilma}                 & ICCV'23    & 74.03     & 89.59     & 93.62       & 66.57    & 63.83     & 80.15     & 85.74      & 38.26     & 61.2     & 81.50      & 88.80      & 48.51\\
                                               & RaSa \cite{bai2023rasa}       & IJCAI'23    & 76.51     & 90.29      & 94.25      & 69.38    & 65.28     & 80.40      & 85.12     & 41.29     & 66.90    & 86.50     & 91.35     & 52.31   \\
                                               & SAP-SAM \cite{wang2024fine}       & MM'24    & 75.05     & 89.93      & 93.73      & -    & 63.97  & 80.84      & 86.17     & -     & 62.85    & 82.65     & 89.85     & -   \\
                                               & IRLT \cite{liu2024causality}       & AAAI'24    & 74.46     & 90.19      & 94.01      & -    & 64.72  & 81.35      & 86.31     & -     & 61.49    & 82.26     & 89.23     & -   \\
                                               & DP \cite{song2024diverse}         & AAAI'24    & 75.66     & 90.59      & 94.07      & 66.58    & 65.61  & \textbf{81.73}      & \textbf{86.95}     & 39.14     & 62.48    & 83.77     & 89.93     & 48.86   \\ 
                                               & UUMSA \cite{zhao2024unifying}         & AAAI'24    & 74.25     & 89.83      & 93.58      & 66.15    & 65.62    & 80.54     & 85.83     & 38.78   & 63.40  & 83.30      & 90.30     & 49.28     \\ 
                                               & PLOT \cite{park2024plot}       & ECCV'24    & 75.28     & 90.42      & 94.12      & -    & 65.76  & 81.39      & 86.73     & -     & 61.80    & 82.85     & 89.45     & -   \\
\hline
                                               & \textbf{AGA(Ours)}        & -       & \textbf{78.36}    & \textbf{91.42}    & \textbf{94.98}    & \textbf{70.72}       & \textbf{67.31} & 81.27 & 85.54 & \textbf{42.31}      & \textbf{67.40} & \textbf{86.94} & \textbf{91.44} & \textbf{52.49}  \\
\hline
\end{tabular}
}
\label{table1}
\end{table*}

\subsection{Datasets and Evaluation Protocol}

We evaluate our method on three popular TBPS datasets, including CUHK-PEDES~\cite{zheng2020dual}, ICFG-PEDES~\cite{ding2021semantically}, and RSTPReid~\cite{zhu2021dssl}. 

\textbf{CUHK-PEDES} is the most commonly used dataset for person search with natural language query. It contains 40,260 pedestrian images and 80,440 texts of 13,003 identities, and each image comes with two hand-annotated descriptions. The dataset is split into three subsets for training, validation, and testing, with non-overlapping identities. Specifically, there are 11,003 identities with 34,054 images and 68108 text descriptions in the training set. The validation set contains 3,078 images and 6,158 text descriptions of 1,000 pedestrians, while the testing set includes 3,074 images with 6,156 captions of another 1,000 pedestrians.

\textbf{ICFG-PEDES} contains 54,522 images of 41,02 pedestrians collected from the MSMT17~\cite{MSMT17}. Each image has one text description. The average description length is 37 words, while the vocabulary contains 5,554 unique words. Compared to CUHK-PEDES, the text description of ICFG-PEDES is more fine-grained and identity-centric. The dataset is divided into training and test sets, containing 34,674 text-image pairs of 3,102 pedestrians and 19,848 text-image pairs of 1,100 people, respectively.

\textbf{RSTPReid} is a recently published dataset also built using MSMT17~\cite{MSMT17} to handle real-world scenarios. It contains 20,505 images of 4,101 identities from 15 cameras in total. Each identity contains five corresponding images from different cameras, each of which is annotated with two textual descriptions. The dataset is split into 3,701 training, 200 validation, and 200 test individuals, respectively.

\textbf{Evaluation Protocol.} To perform a fair comparison with existing methods, all experiments follow the common evaluation settings. For these datasets, the Cumulative Matching Characteristic (CMC) and mean Average Precision (\emph{m}AP) metrics are adopted to evaluate the performance.

\subsection{Implementation Details}
The proposed AGA follows the basic training setting of Rasa~\cite{bai2023rasa}. Specifically, for the text encoder, the maximum length of input tokens is set to 50, and the text Transformer block settings are consistent with Bert~\cite{bert}. While for the image encoder, the images in the dataset are resized to $384\times384$ and the patch size is $16\times16$. Random horizontal flipping, random cropping with padding, and random erasing are employed for image data augmentation. The base learning rate is set to $1 \times 10^{-5}$, and AdamW optimizer~\cite{adamw} is adopted with a weight decay of 0.02. The total number of training epochs is set to 30 for full convergence and the batch size is set to 13. Following the success of Bert~\cite{bert}, the hyper-parameters $\alpha_1$ and $\alpha_2$ are set to 0.05 and 0.15, respectively, which can keep about 15\% of the words selected for masking. We implement our AGA with PyTorch and conduct all experiments on the 4 NVIDIA RTX3090.

\subsection{Comparison with State-of-the-Art Methods}
We compare our AGA with state-of-the-art (SOTA) TBPS approaches on three widely used public datasets. The compared SOTAs include twelve methods without applying Vision-Language Pre-training (VLP) models and four VLP-based models. The comparison results on three benchmark datasets are shown in Table~\ref{table1}, the proposed AGA outperforms existing SOTAs by large margins. Specifically, AGA achieves Rank-1 accuracy of $78.36$\% and \emph{m}AP of $70.72$\% on CUHK-PEDES, significantly improving the Rank-1 accuracy by $1.85$\% and \emph{m}AP by $1.34$\% over the best SOTA RaSa. Similar to the results on the CUHK-PEDES dataset, AGA consistently outperforms current SOTAs on ICFG-PEDES and RSTPReid. Specifically, our AGA improves the Rank-1 accuracy of ICFG-PEDES with $2.03$\% and $0.5$\% on RSTPReid, respectively.
The above results demonstrate the outstanding performance of AGA thanks to its ability in cross-modal 
representation alignment.

\subsection{Ablation Study}
\begin{table}[t]
\renewcommand\arraystretch{1.2}
\setlength{\tabcolsep}{5pt}
\centering
\caption{Ablation study on each component of AGA on CUHK-PEDES.}
\resizebox{0.95\columnwidth}{!}{\begin{tabular}{c|cccc} 
\hline
Module                 & R@1            & R@5            & R@10           & mAP             \\ 
\hline
Baseline              & 76.03          & 90.31          & 93.73          & 69.23       \\
AGM                   & 77.58          & \textbf{91.61}          & 94.89          & 70.50           \\
TEM                   & 76.87          & 90.69          & 94.57          & 69.40           \\
AGM+TEM(AGA)               & \textbf{78.36}          & 91.42         & \textbf{94.98}         & \textbf{70.72}             \\
\hline
\end{tabular}}
\label{ablation_table}
\end{table}

\begin{table}[t]
\renewcommand\arraystretch{1.2}
\setlength{\tabcolsep}{5pt}
\caption{Performance under different attention aggregation strategies on CUHK-PEDES.}
\centering
\resizebox{0.85\columnwidth}{!}{\begin{tabular}{c|cccc} 
\hline
Attention                 & R@1            & R@5            & R@10           & mAP             \\ 
\hline
Mean              & 78.18           & 91.39          & 94.80           & 70.68           \\
Last                   & 78.08          & 91.40          & 94.80          & \textbf{70.75}           \\
Ema                   & \textbf{78.36}          & \textbf{91.42}          & \textbf{94.98}          & 70.72           \\
\hline
\end{tabular}}
\label{att_table}
\end{table}

\begin{figure*}[t]
    \centering
    \includegraphics[width=2.0\columnwidth]{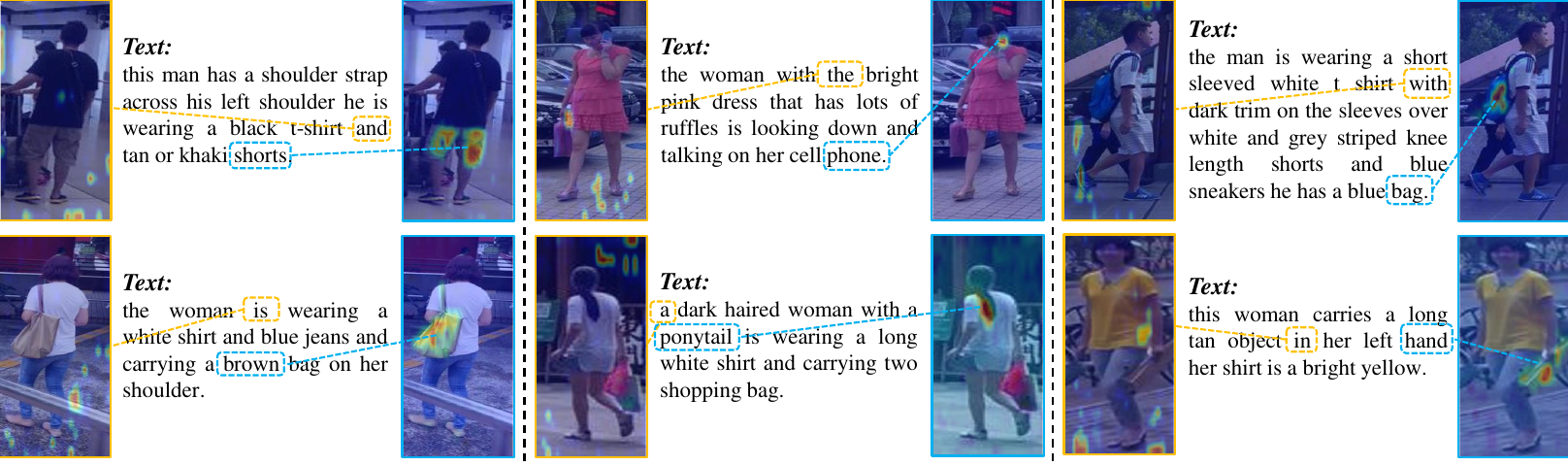}
    \caption{Visualization results of AGM in the cross attention layer. By highly lighting the words with rich semantics, AGM largely improves the quality of the masked text, thereby better aligning cross-modal semantic representations.}
    \label{fig:attvis}
    
\end{figure*}

We conducted the ablation experiment on CUHK-PEDES to evaluate the contribution of each module.
As shown in Table~\ref{ablation_table},
the effectiveness of each module is revealed. Compared with the baseline, AGM improves the Rank-1 accuracy and \emph{m}AP by 1.07\% and 1.12\%, respectively. We also observe 0.36\% Rank-1 accuracy improvements over the baseline when adding TEM.
This implies that enriched text descriptions indeed prevent the loss of VLP's powerful knowledge. 
And when these two modules work together, the Rank-1 accuracy and \emph{m}AP are significantly improved by 1.85\% and 1.34\%. These results demonstrate the effectiveness of each module.

\subsection{Discussions}
\textbf{Effect of attention aggregation strategy.} 
We investigate different attention-acquisition strategies on CUHK-PEDES. After we obtain class attention weights of each layer of the text encoder, there are three different attention aggregation strategies.
Here, 'Mean' aims to directly average the class attention maps of all layers, 'Last' means using only the last layer, while 'Ema' combines the attention maps between each layer in an exponential moving average manner. As shown in Table~\ref{att_table}, when we set $\beta = 0.95$ in Eq.~\eqref{ema}, which means the deep layer has a greater impact on the final attention result, 'Ema' obtains the best performance and works better than the other two strategies.


\begin{table}[t]
\renewcommand\arraystretch{1.2}
\setlength{\tabcolsep}{5pt}
\centering
\caption{Performance under different probability $p_{tem}$ on CUHK-PEDES.}
\resizebox{0.8\columnwidth}{!}{
\begin{tabular}{c|cccc} 
\hline
$p_{tem}$                 & R@1            & R@5            & R@10           & mAP             \\ 
\hline
0.1               & 78.23          & 91.14          & 94.83          & 70.65           \\
0.2               & \textbf{78.39}          & 91.27          & 94.72          & 70.71           \\
0.3               & 78.36          & \textbf{91.42}          & \textbf{94.98}          & \textbf{70.72}           \\
0.4               & 78.11          & 91.48          & 94.76          & 70.63           \\
\hline
\end{tabular}}
\label{p_table}
\end{table}

\noindent\textbf{Impact of the hyperparameter $p_{tem}$}.
In Table~\ref{p_table}, we investigate the impact of probability $p_{tem}$, which determines whether the new text replaces the original text. Small $p_{tem}$ means that few descriptions are replaced, which limits the creation of new 'views' on the text.
When $p_{tem}$ increases, the original sentence is replaced with the new sentence obtained by TEM with a greater probability. 
While $p_{tem}$ is set to 0.3, we achieve the best results in our approach with the performance of 78.36\%, 91.42\%, 94.98\%, 70.72\% on R@1, R@5, R@10, \emph{m}AP, respectively.

\noindent\textbf{Impact of the hyperparameter $\alpha_1$}.
From Table~\ref{table:toy}, we can observe that there are still around 8.3\% of meaningless words remaining in the AGA. Therefore, in this part, we explore whether further reducing the proportion of meaningless words is useful. In Eq.~\ref{prob}, we set a hyper-parameter $\alpha_1$ to control the lower bound of masking probability to each text. Hence, we conduct experiments under different $\alpha_1$ and show the result in Table~\ref{table:alpha1}. 
Intuitively, totally ignoring all the meaningless words may work better.
However, as shown in Table~\ref{table:alpha1}, the optimal performance reaches when $\alpha_1 = 0.05$. We suppose that the smaller parameters cause the network to overly focus on local regions of the text, while this reduces its understanding of the overall semantics.

\begin{table}[t]
\renewcommand\arraystretch{1.2}
\setlength{\tabcolsep}{5pt}
\centering
\caption{Performance under a different number of candidate words $k$. In the TEM, only the top-$k$ corresponding word will be considered to replace the original word.}
\resizebox{0.85\columnwidth}{!}{
\begin{tabular}{c|cccc} 
\hline
Setting              & R@1            & R@5            & R@10           & mAP             \\ 
\hline
$k = 3$               & 78.00          & 91.06          & 94.73          & 70.52           \\
$k = 5$               & \textbf{78.36}          & \textbf{91.42}          & \textbf{94.98}          & \textbf{70.72}           \\
$k = 10$              & 78.05          & 91.26          & 94.86          & 70.61          \\
\hline
\end{tabular}}
\label{topk_table}
\end{table}

\noindent\textbf{Number of candidate words $k$.} In~\ref{topk_table}, we evaluate the number of candidate words $k$ used to replace the original word. Here, $k$ means getting the top-k corresponding words that refer to the logit, and then selecting a word among them to replace the original word that uses the logit as the distribution function. 
When $k$ is small, it is insufficient to generate multi-views, making it difficult to guide the model to learn better representations.
However, setting a larger $k$, makes it difficult to obtain so many words with similar semantics in the vocabulary, leading to inappropriate replacement. 
Overall, our method achieves the best performance with Rank-1 accuracy and \emph{m}AP when $k$ is set to 5.

\begin{table}[t]
\renewcommand\arraystretch{1.2}
\setlength{\tabcolsep}{5pt}
\centering
\caption{The influence of the lower bound of the masking probability $\alpha_1$ on CUHK-PEDES. $Ratio_{v}$ is defined as $N_{vacuous}/N_{Masked}$. Although lower $\alpha_1$ further reduces the probability of masking meaningless words, they can lead to overfitting of the network to a small number of words, thereby reducing global understanding.}
\resizebox{0.9\columnwidth}{!}
{\begin{tabular}{c|c|cccc} 
\hline
Setting      &$Ratio_{v}$     & R@1            & R@5            & R@10           & mAP             \\ 
\hline
$\alpha_1 = -0.05$        & $\approx$0\%           & 76.03          & 90.31          & 93.73          & 69.23       \\
$\alpha_1 = 0.0$        & $\approx$7\%           & 75.50          & 89.24          & 93.63          & 69.15       \\
$\alpha_1 = 0.05$    &$\approx$8\%            & 76.51          & 90.29          & 94.25          & 69.38           \\
$\alpha_1 = 0.1$      &$>10$\%  & \textbf{77.58} & \textbf{91.61} & \textbf{94.89} & \textbf{70.50}  \\
\hline
\end{tabular}}
\label{table:alpha1}
\end{table}

\begin{table}[t]
\renewcommand\arraystretch{1.2}
\setlength{\tabcolsep}{5pt}
\centering
\caption{The influence of the hyper-parameter $\beta$ of EMA on CUHK-PEDES. The optimal performance reaches when $\beta$ is set to 0.95, which indicates the attention results from deep layers can better reflect the weight of the text.}
\resizebox{0.85\columnwidth}{!}{
\begin{tabular}{c|cccc} 
\hline
Setting           & R@1            & R@5            & R@10           & mAP             \\ 
\hline
$\beta = 0.90$               & 78.21           & 91.34          & 94.72           & 70.66           \\
$\beta = 0.95$               & \textbf{78.36}          & 91.42          & \textbf{94.98}          & \textbf{70.72}           \\
$\beta = 0.99$               & 77.95           & \textbf{91.47}          & 94.68           & 70.59          \\
\hline
\end{tabular}}
\label{beta}
\end{table}

\begin{figure}[t]
    \centering
    \includegraphics[width=1.0\columnwidth]{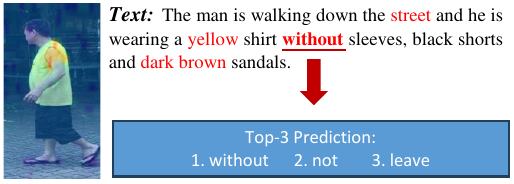}
    \caption{An example of AGA. We mark the Top-5 words with the highest attention weight in the text in red and show the prediction when masking the word "without". Therefore, besides simple nouns, AGA can also focus on the important correlations in the text.}
    \label{fig:vis2}
\end{figure}

\noindent\textbf{Visualization.} 
In Figure~\ref{fig:attvis}, we visualized the cross-attention maps between different masked text tokens and image tokens.
Compared to the random mask strategy, which contains a large number of meaningless words that can not bring efficient interaction between the two modalities, AGM highlights the words with more semantic information, largely improving the quality of masked words. 
Therefore, the model can capture the corresponding semantic information from images during cross-attention, thereby showing effective interaction and further boosting performance. Meanwhile, we further visualized an example of how AGA works. Although examples in the paper always focus on a specific object (e.g. phone, bag), as shown in Figure~\ref{fig:vis2}, those words that describe the correlation also show high masking probability in AGA. An interesting observation shown in Figure~\ref{fig:vis2} is that the network always considers "with" as a "semantically vacuous" word while considering "without" as a "meaningful" word. It demonstrates that the AGA method proposed in this paper is not limited to the understanding of simple nouns in meaningful vocabulary, it also encompasses the different important relationships described within the context. The ability of AGA to interpret complex relational structures between elements in sentences allows it to form a more comprehensive understanding of textual descriptions and thus encouraging more directional masking strategy. This facilitates more effective semantic parsing, enhancing the model's capability to extract the descriptions, thus providing a more robust representation to match visual content.

\section{Conclusion}
In this paper, we propose an Attention-Guided Alignment (AGA) framework for text-based person search (TBPS) to pursue a more stable and efficient interactive training between the text descriptions and image samples. 
To overcome inefficient training in the random-based mask modeling caused by those semantically vacuous words, we introduce Attention-Guided Mask (AGM) Modeling.
AGM dynamically masks semantically meaningful words in a resource-efficient manner by aggregating the class attention weight of the text encoder, thereby enhancing cross-modal semantic alignment.
Meanwhile, by observing the inaccurate and monotonous descriptions in the TBPS datasets which potentially undermine the rich representational power of VLP models and lead the training to overfitting, we further introduce a Text Enrichment Module (TEM).    
The TEM module utilizes logit from the MLM head to perform synonym replacement, which largely enriches the text description with minimal semantic deviation.
Extensive experiments across various TBPS benchmark datasets validate the superiority and effectiveness of our proposed AGA framework.

\section*{Acknowledgments}
This work was supported by the National Key R\&D Program of China (No. 2022ZD0118202), the National Science Fund for Distinguished Young Scholars (No.62025603), the National Natural Science Foundation of China (No. U21B2037, No. U22B2051, No. 62176222, No. 62176223, No. 62176226, No. 62072386, No. 62072387, No. 62072389, No. 62002305, and No. 62272401), and the Natural Science Foundation of Fujian Province of China (No. 2021J01002, No. 2022J06001).

\bibliography{refer}
\bibliographystyle{IEEEtran}

\vfill

\end{document}